\documentclass[letterpaper]{article} 
\usepackage{aaai24}  
\usepackage{times}  
\usepackage{helvet}  
\usepackage{courier}  
\usepackage[hyphens]{url}  
\usepackage{graphicx} 
\usepackage{natbib}  
\usepackage{caption} 
\usepackage{algorithm}
\usepackage{algorithmic}

\usepackage{subcaption}
\usepackage{booktabs}
\usepackage{multirow}
\usepackage{color}
\usepackage{amssymb}
\usepackage{amsmath}

\usepackage{newfloat}
\usepackage{listings}
\DeclareCaptionStyle{ruled}{labelfont=normalfont,labelsep=colon,strut=off} 
\floatstyle{ruled}
\newfloat{listing}{tb}{lst}{}
\floatname{listing}{Listing}
\title{ODTrack: Online Dense Temporal Token Learning for Visual Tracking}

\author{
    Yaozong Zheng\textsuperscript{\rm 1,2}, Bineng Zhong\textsuperscript{\rm 1,2}\thanks{Corresponding author.}, Qihua Liang\textsuperscript{\rm 1,2}, Zhiyi Mo\textsuperscript{\rm 3}, Shengping Zhang\textsuperscript{\rm 4}, Xianxian Li\textsuperscript{\rm 1,2}
}
\affiliations{
    \textsuperscript{\rm 1}Key Laboratory of Education Blockchain and Intelligent Technology, Ministry of Education, Guangxi Normal University\\
    \textsuperscript{\rm 2}Guangxi Key Lab of Multi-Source Information Mining \& Security, Guangxi Normal University\\
    \textsuperscript{\rm 3}Guangxi Key Laboratory of Machine Vision and Intelligent Control, Wuzhou University
    \textsuperscript{\rm 4}Harbin Institute of Technology\\

    yaozongzheng@stu.gxnu.edu.cn, bnzhong@gxnu.edu.cn, qhliang@gxnu.edu.cn\\
    zhiyim@gxuwz.edu.cn, s.zhang@hit.edu.cn, lixx@gxnu.edu.cn\\
}

\usepackage{bibentry}

\begin{document}

\maketitle

\begin{abstract}


Online contextual reasoning and association across consecutive video frames are critical to perceive instances in visual tracking. However, most current top-performing trackers persistently lean on sparse temporal relationships between reference and search frames via an offline mode. Consequently, they can only interact independently within each image-pair and establish limited temporal correlations. To alleviate the above problem, we propose a simple, flexible and effective video-level tracking pipeline, named \textbf{ODTrack}, which densely associates the contextual relationships of video frames in an online token propagation manner. ODTrack receives video frames of arbitrary length to capture the spatio-temporal trajectory relationships of an instance, and compresses the discrimination features (localization information) of a target into a token sequence to achieve frame-to-frame association. This new solution brings the following benefits: 1) the purified token sequences can serve as prompts for the inference in the next video frame, whereby past information is leveraged to guide future inference; 2) the complex online update strategies are effectively avoided by the iterative propagation of token sequences, and thus we can achieve more efficient model representation and computation. ODTrack achieves a new \textit{SOTA} performance on seven benchmarks, while running at real-time speed.
Code and models are available at \url{https://github.com/GXNU-ZhongLab/ODTrack}.




\end{abstract}

\section{Introduction}

Visual tracking aims to uniquely identify and track an object within a video sequence by using arbitrary target queries.
In the visual world, objects rarely exist in isolation but rather within a larger and dynamic context. Therefore, visual perception is a complex process that involves interpreting and understanding the surrounding environment of an object. 
In such a situation, equipping a model with the ability to perform online contextual reasoning and establish associations presents a challenge in the field of visual tracking.

   \begin{figure}[t]
      \centering
      \includegraphics[width=0.85\columnwidth]{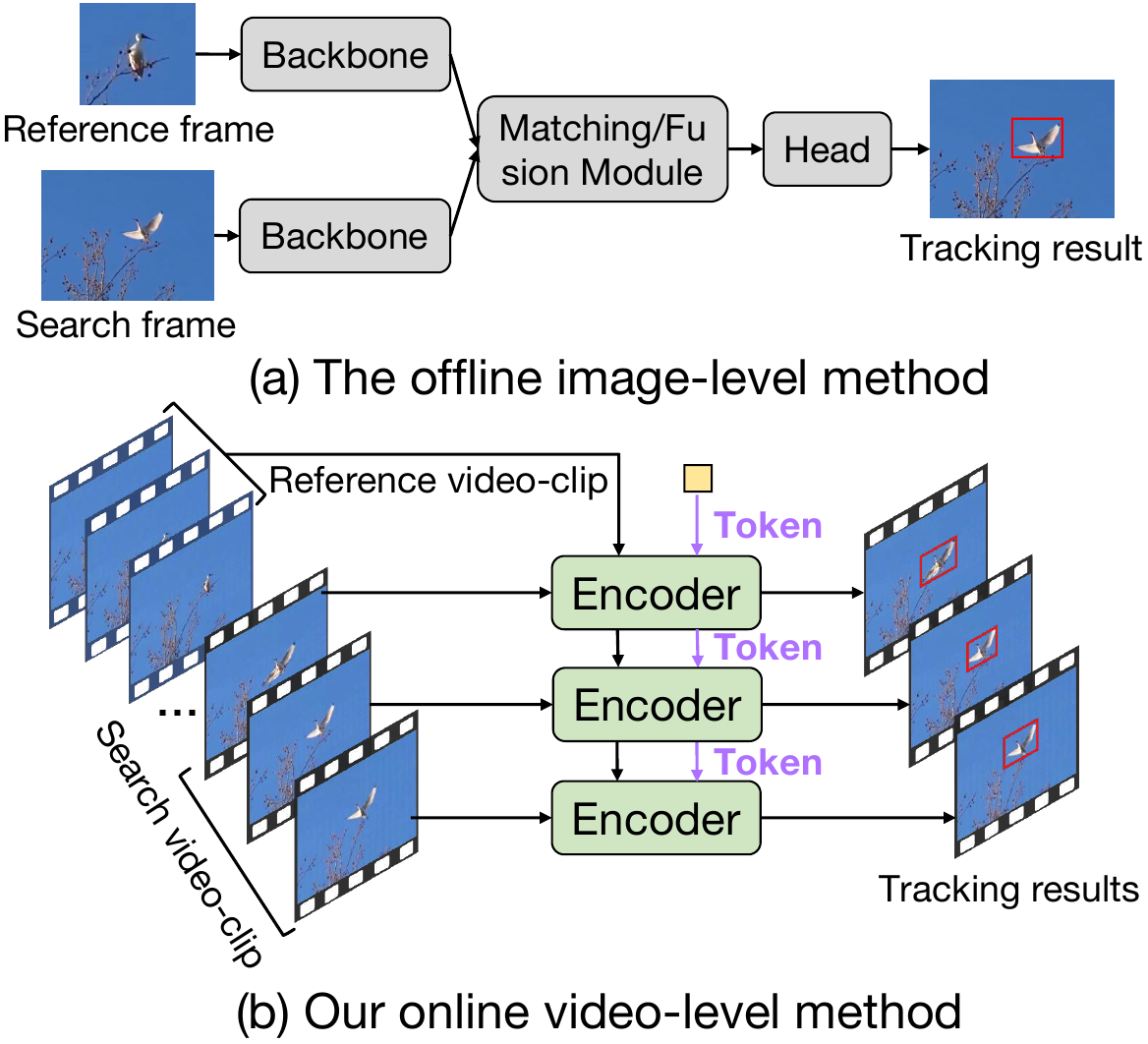}
       \caption{Comparison of tracking methods. (a) The offline image level tracking methods\cite{SiamRPN++,transt} based on sparse sampling and image-pair matching. (b) Our online video-level tracking method based on video sequence sampling and temporal token propagation.}
       \label{fig:motivation}
    \end{figure}

Despite this challenge, a significant number of current tracking methods overlook this problem and instead rely on the offline image-pair matching to localize instances in the current frame.
As shown in Fig.\ref{fig:motivation}(a), these offline methods\cite{SiamFC,SiamRPN++,transt,stark,ostrack,mixformer} typically follow a three-phase process: (\textit{i}) extracting features by sampling two video frames (i.e., reference and search frames); (\textit{ii}) propagating the initial target information from the reference to the search frame through a matching/fusion module; and (\textit{iii}) utilizing a bounding box prediction head to output the localization results.
Most trackers have performed well under this paradigm, but still exhibit the following drawbacks: (1) \textit{The sampling frames are sparse} (i.e., using only one reference frame and one search frame). Although visual tracking inherently contains rich temporal data, this simple sampling strategy falls short in accurately representing the motion state of an object, posing a significant challenge for trackers to comprehend dynamic video content, and (2) \textit{The target information is matched offline and limited to image-pair level}, preventing the association of the targets across video frames. Traditional feature matching/fusion methods\cite{Siamban,Ocean,SiamGAT,SBT} focus on the appearance similarity of object, without considering the property that tracking instance rely on continuous cross-frame associations.


To incorporate temporal information into the model, some approaches commonly design online updating techniques, such as updating templates\cite{stark,mixformer} and updating model parameters\cite{DiMP50}.
Despite being successful, these methods still rely on sparse sampling frames (i.e., reference, search, and update frames) and do not effectively explore how information is propagated online across search frames.
This inspired us to think: 
\textit{can our visual tracking algorithm densely associate and perceive an object in a video streaming context?}

The answer is affirmative. Unlike conventional approaches that rely on offline image-pair matching with sparse sampling frames, this paper proposes \textbf{ODTrack}, a novel video-level framework for visual tracking that capitalizes on video stream modeling.
Specifically, we reformulate object tracking as a token sequence propagation task that densely associates the contextual relationships of across video frames in an auto-regressive manner, as shown in Fig.\ref{fig:motivation}(b). To overcome the limitations of traditional image-pair sampling strategy and explore the rich temporal dependencies, we extend the model’s input from image-pair to the level of a video stream.
Under this new input paradigm, we design two simple yet effective temporal token propagation attention mechanism that captures the spatio-temporal trajectory relationships of the target instance using an online token propagation manner, thus allowing the processing of video-level inputs of arbitrary length.
Notably, we treat each video sequence as a continuous sentence, enabling us to employ language modeling for a comprehensive contextual understanding of the video content. This novel approach significantly distinguishes our tracker from traditional methods \cite{stark,ostrack,mixformer} and greatly strengthens its ability to understand the spatio-temporal trajectory of target instance. 

The main contributions of this work are as follows.
    \begin{itemize}
    \item We propose a novel video-level tracking pipeline, called ODTrack. In contrast to existing tracking approaches based on sparse temporal modeling, we employ a token sequence propagation paradigm to densely associate contextual relationships across video frames.
    
    
    

    \item We introduce two temporal token propagation attention mechanisms that compress the discriminative features of the target into a token sequence. This token sequence serves as a prompt to guide the inference of future frames, thus avoiding complex online update strategies.
    


    
    \item Our approach achieves a new state-of-the-art tracking results on seven visual tracking benchmarks, including LaSOT, TrackingNet, GOT10K, LaSOT$_{\rm{ext}}$, VOT2020, TNL2K, and OTB100.
    

    
    \end{itemize}


\section{Related Work}

\subsubsection{Traditional Tracking Framework.} The current popular trackers\cite{SiamFC,SiamRPN++,transt,ostrack} are dominated by the Siamese tracking paradigm, which achieves tracking by image-pair matching. 
To improve the accuracy and robustness of trackers, several different approaches are proposed, such as prediction head networks \cite{SiamRPN,Siamban,Ocean}, cross-correlation modules \cite{ACM,PG-Net,transt}, powerful backbone \cite{simtrack,mixformer} and attention mechanisms \cite{SiamGAT,siamattn}. 
In recent years, the introduction of the transformer \cite{attention} enables trackers \cite{stark,SBT,mixformer,ostrack} to explore more powerful and deeper feature interactions, resulting in significant advances in tracking algorithm development.
However, most of these methods are designed based on offline mode and sparse image-pair strategy.
With this design paradigm, the tracker struggles to accurately comprehend the object’s motion state in the temporal dimension and can only resort to traditional Siamese similarity for appearance modeling.
In contrast to these approaches, we reformulate object tracking as a token sequence propagation task and aim to extend Siamese tracker to efficiently exploit target temporal information in an auto-regressive manner.

\subsubsection{Temporal Modelling in Visual Tracking.} 
Multi-object tracking algorithms\cite{trackformer,motr} typically involve the recognition and association of individual objects in a video, making the study of trajectory information a common practice. However, there exists a relatively limited amount of research exploring the utilization of spatio-temporal trajectory information in single-object tracking algorithms.

To explore temporal cues within the Siamese framework, several online update methods are carefully designed.
UpdateNet\cite{updateNet} introduces an adaptive updating strategy, which utilizes a custom network to fuse accumulated templates and generate a weighted updated template feature for visual tracking. DCF-based trackers\cite{atom,DiMP50,PrDiMP-50} excel at updating model parameters online using sophisticated optimization techniques, thereby improving the robustness of the tracker. STMTrack\cite{STMTrack} and TrDiMP\cite{trdimp} employ attention mechanism to effectively extract contextual information along the temporal dimension. STARK\cite{stark} and Mixformer\cite{mixformer} specifically design target quality branch for updating template frame, which aids in improving the tracking results.
Recently, there has been a gradual surge in research attention towards modeling temporal context from various perspectives.
TCTrack \cite{TCTrack} introduces an online temporal adaptive convolution and an adaptive temporal transformer that aggregates temporal contexts at two levels containing feature extraction and similarity map refinement. VideoTrack \cite{VideoTrack} designs a new tracker based on video transformer and uses a simple feedforward network to encode temporal dependencies.
ARTrack \cite{ARTrack} presents a new time-autoregressive tracker that estimates the coordinate sequence of an object progressively.

   \begin{figure*}
      \centering
      \includegraphics[width=0.81\linewidth]{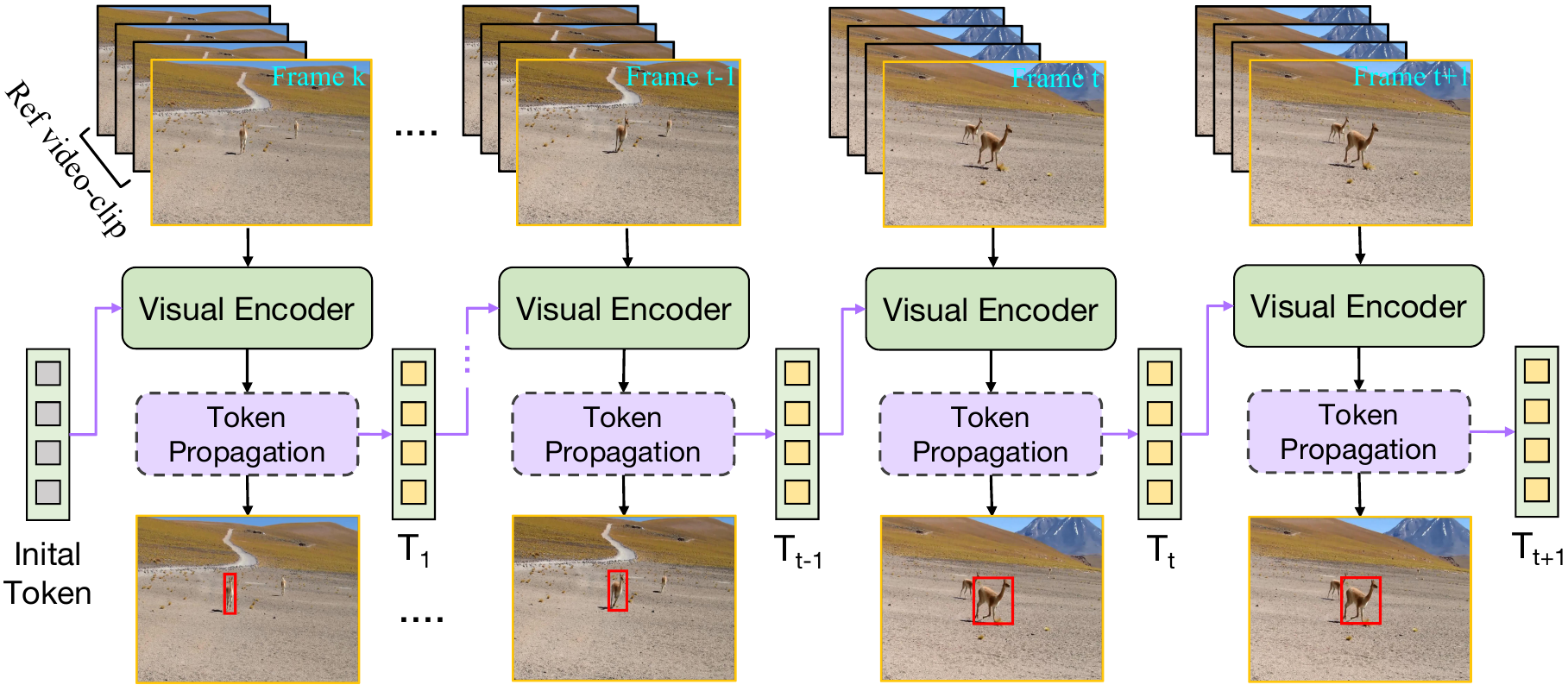}
       \caption{\textbf{ODTrack Framework Architecture}.
       The ODTrack pipeline takes video clips, consisting of reference and search frames, of arbitrary length as input. Then, the model utilizes a temporal token propagation attention mechanism to generate a temporal token for each video frame. These temporal tokens are subsequently propagated to the following frames in an auto-regressive manner, enabling cross-frame propagation of target trajectory information.
       }
       \label{fig:framework}
    \end{figure*}

Nevertheless, the above tracking algorithms still suffer from the following limitations: (1) The optimization process is complex, involving the design of specialized loss functions\cite{DiMP50}, multi-stage training strategies\cite{stark}, and manual update rules\cite{stark}, and (2) Although they explore temporal information to some extent, they fail to investigate \textit{how temporal cues propagate across search frames}.
In this work, we introduce a new dense context propagation mechanism from a token propagation perspective, which offers a solution to circumvent intricate optimization processes and training strategies. Further, we propose a new baseline approach, called ODTrack, focused on unlocking the potential of temporal modeling through the propagation of target motion/trajectory information.

\section{Approach}

We introduce ODTrack, a new video-level framework that employs token sequence propagation for visual tracking, as shown in Fig.\ref{fig:framework}.
This section first describes the concept of video-level visual object tracking, followed by the introduction of temporal token propagation attention mechanism and how they are trained in a new design paradigm.

\subsection{Question Formulation}

To provide a comprehensive understanding of our ODTrack framework, it is pertinent to first offer a review of previously prominent image-pair matching tracking methodologies\cite{SiamFC,transt,ostrack}.

Given a pair of video frames, i.e., a reference frame $R \in \mathbb{R}^{3 \times H_r \times W_r}$ and a search frame $S \in \mathbb{R}^{3 \times H_s \times W_s}$, the mainstream visual trackers $\Psi$ are formulated as $B \gets \Psi:\{R, S\}$, where $B$ denotes the predicted box coordinates of the current search frame.
If $\Psi$ is a conventional convolutional siamese tracker\cite{SiamRPN++,Siamban,transt}, it undergoes three stages, namely feature extraction, feature fusion, and bounding box prediction. Whereas if $\Psi$ is a transformer tracker\cite{ostrack,mixformer,simtrack}, it consists solely of a backbone and a prediction head network, where the backbone integrates the processes of feature extraction and fusion.

Specifically, the transformer tracker receives a series of non-overlapping image patches (the resolution of each image patch is $p \times p$) as input. This means that a 2D reference-search image pair needs to pass through a patch embedding layer to generate multiple 1D image token sequences $\{f_r \in \mathbb{R}^{D \times N_r}, f_s \in \mathbb{R}^{D \times N_s} \}$, where $D$ is the token dimension, $N_r = H_rW_r / p^2$, and $N_s = H_sW_s / p^2$. These 1D image tokens are then concatenated and loaded into a $L$-layer transformer encoder for feature extraction and relationship modeling. 
Each transformer layer $\delta$ contains a multi-head attention and a multi-layer perceptron.
Here, we formulate the forward process of the $l^{th}$ transformer layer as follows:

   \begin{equation}
     f_{rs}^{l} = \delta^{l}(f_{rs}^{l-1}), l=1,2,...,L
     \label{eq:backbone}
   \end{equation}
where $f_{rs}^{l-1}$ denotes the concatenated token sequence of the reference-search image pair generated from the $(l-1)^{th}$ transformer layer, and $f_{rs}^{l}$ represents the token sequence generated by the current $l^{th}$ transformer layer.

By adopting the modeling approach mentioned above, we can construct a concise and elegant tracker to achieve per-frame tracking. Nevertheless, this modeling approach has a clear drawback. The created tracker solely focuses on intra-frame target matching and lacks the ability to establish inter-frame associations necessary for tracking object across a video stream. Consequently, this limitation hinders the research of video-level tracking algorithms.

In this work, we aim to alleviate this challenge and propose a new design paradigm for video-level tracking algorithms.
First, we extend the inputs of the tracking framework from the image-pair level to the video level for temporal modeling. Then, we introduce a new temporal token/prompt $T$ designed to propagate information about the appearance, spatio-temporal location and trajectory of the target instance in a video sequence. Formally, we formulate video-level tracking as follows:
   \begin{equation}
     B \gets \Psi:\{R_1, R_2,..., R_k, S_1, S_2,..., S_n, T\}
     \label{eq:tracker}
   \end{equation}
where $\{R_1, R_2,..., R_k\}$ denotes the reference frames of length $k$, and $\{S_1, S_2,..., S_n\}$ represents the search frames of length $n$. Our video-level tracking framework receives video clip of arbitrary length to model spatio-temporal trajectory relationships of the target object. We describe the proposed core module in more detail in the next section.

\subsection{Video-Level Tracking Pipeline}

Fig.\ref{fig:framework} gives an overview of our ODTrack framework. In this section, our focus lies in constructing a video-level tracking pipeline.
Theoretically, we model the entire video as a continuous sequence, and decode the localization of target frame by frame in an auto-regressive manner.
Firstly, we present a novel video sequence sampling strategy designed specifically to meet the input requirements of the video-level model. Subsequently, to capture the spatio-temporal trajectory information of the target instance within the video sequences, we introduce two simple yet effective temporal token propagation attention mechanisms.

\subsubsection{Video Sequence Sampling Strategy}
Most existing trackers \cite{stark,mixformer,ostrack} commonly sample image-pairs within a short-term interval, such as 50, 100, or 200 frame intervals. However, this sampling approach poses a potential limitation as these trackers fail to capture the long-term motion variations of the tracked object, thereby constraining the robustness of tracking algorithms in long-term scenarios. 

To obtain richer spatio-temporal trajectory information of the target instance from long-term video sequences, we deviate from the traditional short-term image-pair sampling method and propose a new video sequence sampling strategy.
Specifically, we establish a larger sampling interval and randomly extract multiple video frames within this interval to form video clips $\{R_1, R_2,..., R_k, S_1, S_2,..., S_n\}$ of any lengths. Although this sampling approach may seem simplistic, it enables us to approximate the content of the entire video sequence. This is crucial for video-level modeling.

\subsubsection{Temporal Token Propagation Attention Mechanism}
Instead of employing a complex video transformer \cite{VideoTrack} as the foundational framework for encoding video content, we approach the design from a new perspective by utilizing a simple 2D transformer architecture, i.e., 2D ViT \cite{vit}.

\begin{figure}[t]
      \centering
      \includegraphics[width=0.96\linewidth]{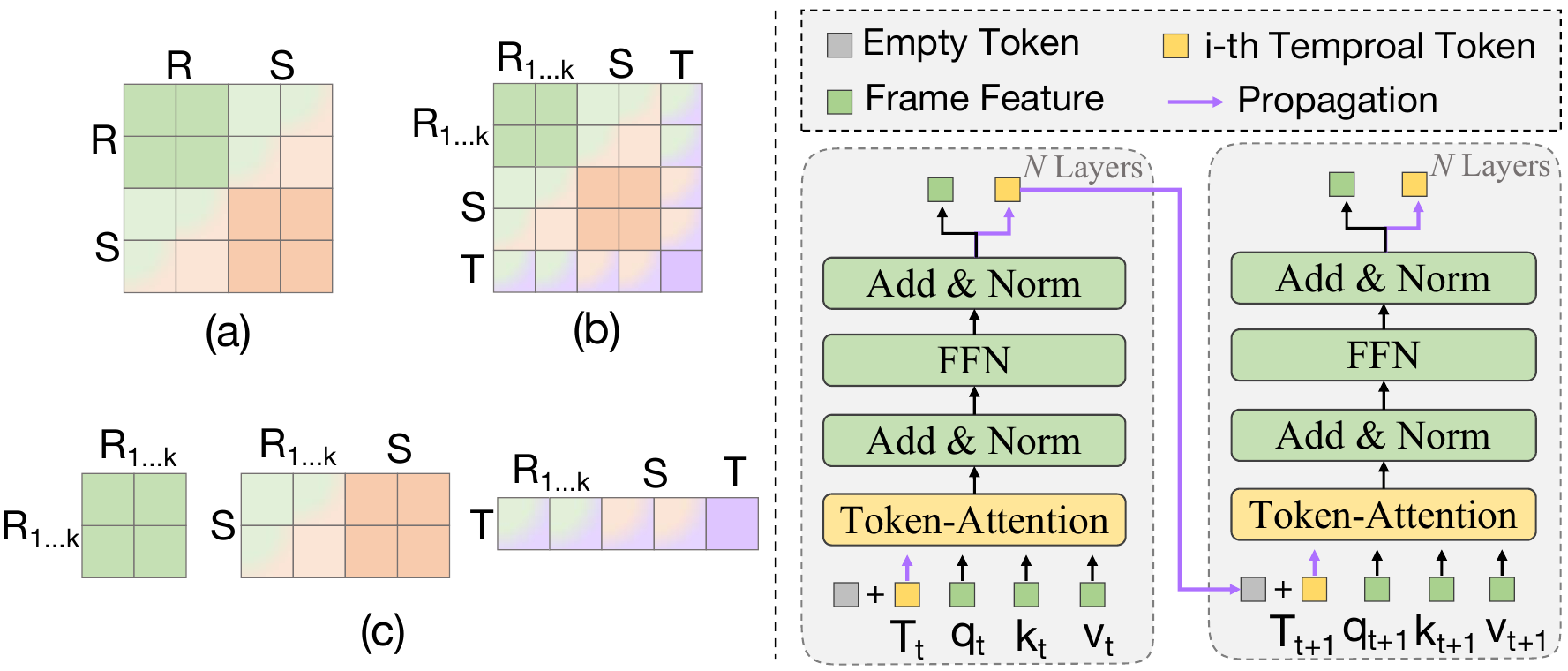}
       \caption{Left: the architecture of temporal token propagation attention mechanism. Right: illustration of online token propagation. (a) Original reference-search attention mechanism, (b) and (c) Different variants of the proposed temporal token propagation attention mechanisms. $R$ is a single reference frame, $R_{1...k}$ denotes  the reference frames of length $k$, $S$ represents the current search frame, and $T$ is the temporal token sequence of current video frames.
       }
       \label{fig:module}
    \end{figure}

To construct an elegant instance-level inter-frame correlation mechanism, it is imperative to extend the original 2D attention operations to extract and integrate video-level features.
In our approach, we design two temporal token attention mechanisms based on the concept of \textbf{\textit{compression-propagation}}, namely \textit{concatenated token attention mechanism} and \textit{separated token attention mechanism}, as shown in Fig.\ref{fig:module}(left).
The core design involves injecting additional information into the attention operations, such as more video sequence content and temporal token vectors, enabling them to extract richer spatio-temporal trajectory information of the target instance.

In Fig.\ref{fig:module}(a), the original attention operation commonly employs an image pair as inputs, where the process of modeling their relationships can be represented as $f=\textnormal{Attn}([R, S])$. In this paradigm, the tracker can only engage in independent interactions within each image pair, establishing limited temporal correlations. 
In Fig.\ref{fig:module}(b), the proposed concatenated token attention mechanism extends the input to the aforementioned video sequence, enabling dense modeling of spatio-temporal relationships across frames.
Inspired by the contextual nature of language formed through concatenation, we apply the concatenation operation to establish context for video sequences as well.
Its formula can be represented as:
   \begin{equation}
     \begin{split}
        f_t &= \textnormal{Attn}([R_1, R_2,..., R_k, S_t, T_t]) \\
        &= \sum_{s''t''} V_{s''t''} \cdot \frac{\exp \langle q_{st}, k_{s''t''} \rangle}{\sum_{s't'} \exp \langle q_{st}, k_{s't'} \rangle}
      \end{split}
     \label{eq:msa}
   \end{equation}
Where $T_t$ is the temporal token sequence of $t^{th}$ video frame. $[\cdots,\cdots]$ denotes concatenation among tokens. $q_{st}$, $k_{st}$ and $v_{st}$ are spatio-temporal linear projections of the concatenated feature tokens.

It is worth noting that we introduce a temporal token for each video frame, with the aim of storing the target trajectory information of the sampled video sequence. In other words, we \textbf{\textit{compress}} the current spatio-temporal trajectory information of the target into a token vector, which is used to \textbf{\textit{propagate}} to the subsequent video frames.

Once the target information is extracted by the temporal token, we propagate the token vector from $t^{th}$ frame to $(t+1)^{th}$ frame in an auto-regressive manner, as shown in Fig.\ref{fig:module}(right).
Firstly, the $t^{th}$ temporal token $T_{t}$ is added to the $(t+1)^{th}$ empty token $T_{empty}$, resulting in an update of the content token $T_{t+1}$ for $(t+1)^{th}$ frame, which is then propagated as input to the subsequent frames. Formally, the propagation process is:
   \begin{equation}
     \begin{split}
        T_{t+1} &= T_t + T_{empty} \\
        f_{t+1} &= \textnormal{Attn}([R_1, R_2,..., R_k, S_{t+1}, T_{t+1}])
      \end{split}
     \label{eq:propagate}
   \end{equation}

In this new design paradigm, we can employ temporal tokens as prompts for inferring the next frame, leveraging past information to guide future inference. Moreover, our model implicitly propagates appearance, localization, and trajectory information of the target instance through online token propagation. This significantly improves tracking performance of the video-level framework.

On the other hand, as illustrated in Fig.\ref{fig:module}(c), the proposed separated token attention mechanism decomposes attention operation into three sub-processes: self-information aggregation between reference frames, cross-information aggregation between reference and search frames, and cross-information aggregation between temporal token and video sequences. This decomposition improves the computational efficiency of the model to a certain extent, while the token propagation aligns with the aforementioned procedures.

\subsubsection{Discussions with Online Update.}

Most previous tracking algorithms combine online updating methods to train a spatio-temporal tracking model, such as adding an extra score quality branch\cite{stark} or an IoU prediction branch\cite{atom}. They typically require complex optimization processes and update decision rules. In contrast to these methods, we avoid complex online update strategies by utilizing online iterative propagation of token sequences, enabling us to achieve more efficient model representation and computation.

\subsection{Prediction Head and Loss Function}

For the design of the prediction head network, we employ conventional classification head and bounding box regression head to achieve the desired outcome.
The classification score map $\mathbb{R}^{1 \times \frac{H_s}{p} \times \frac{W_s}{p}}$, bounding box size $\mathbb{R}^{2 \times \frac{H_s}{p} \times \frac{W_s}{p}}$, and offset size $\mathbb{R}^{2 \times \frac{H_s}{p} \times \frac{W_s}{p}}$ for the prediction are obtained through three sub-convolutional networks, respectively. 
We adopt the focal loss\cite{focalloss} as classification loss $L_{cls}$, and the $L_1$ loss and $GIoU$ loss\cite{giou} as regression loss. The total loss $L$ can be formulated as:
   \begin{equation}
      L = L_{cls} + \lambda_{1}L_{1} + \lambda_{2}L_{GIoU}
     \label{eq:loss}
   \end{equation}
where $\lambda_{1}$ = 5 and $\lambda_{2}$ = 2 are the regularization parameters. Since we use video segments for modeling, the task loss is computed independently for each video frame, and the final loss is averaged over the length of the search frames.

\begin{table}[t]
\centering
\caption{Comparison of model parameters, FLOPs, and inference speed.}
\resizebox{\linewidth}{!}{
\begin{tabular}{l|cccccc}
\toprule
Method & Type & Resolution & Params & FLOPs & Speed  & Device \\
\midrule
SeqTrack & ViT-B & $384\times384$ & 89M & 148G & 11$fps$ & 2080Ti \\
ODTrack & ViT-B & $384\times384$ & 92M & 73G & 32$fps$ & 2080Ti \\
\bottomrule
\end{tabular} }
\label{tab:param}
\end{table}

\section{Experiments}

   \begin{figure}[t]
      \centering
      \includegraphics[width=0.9\linewidth]{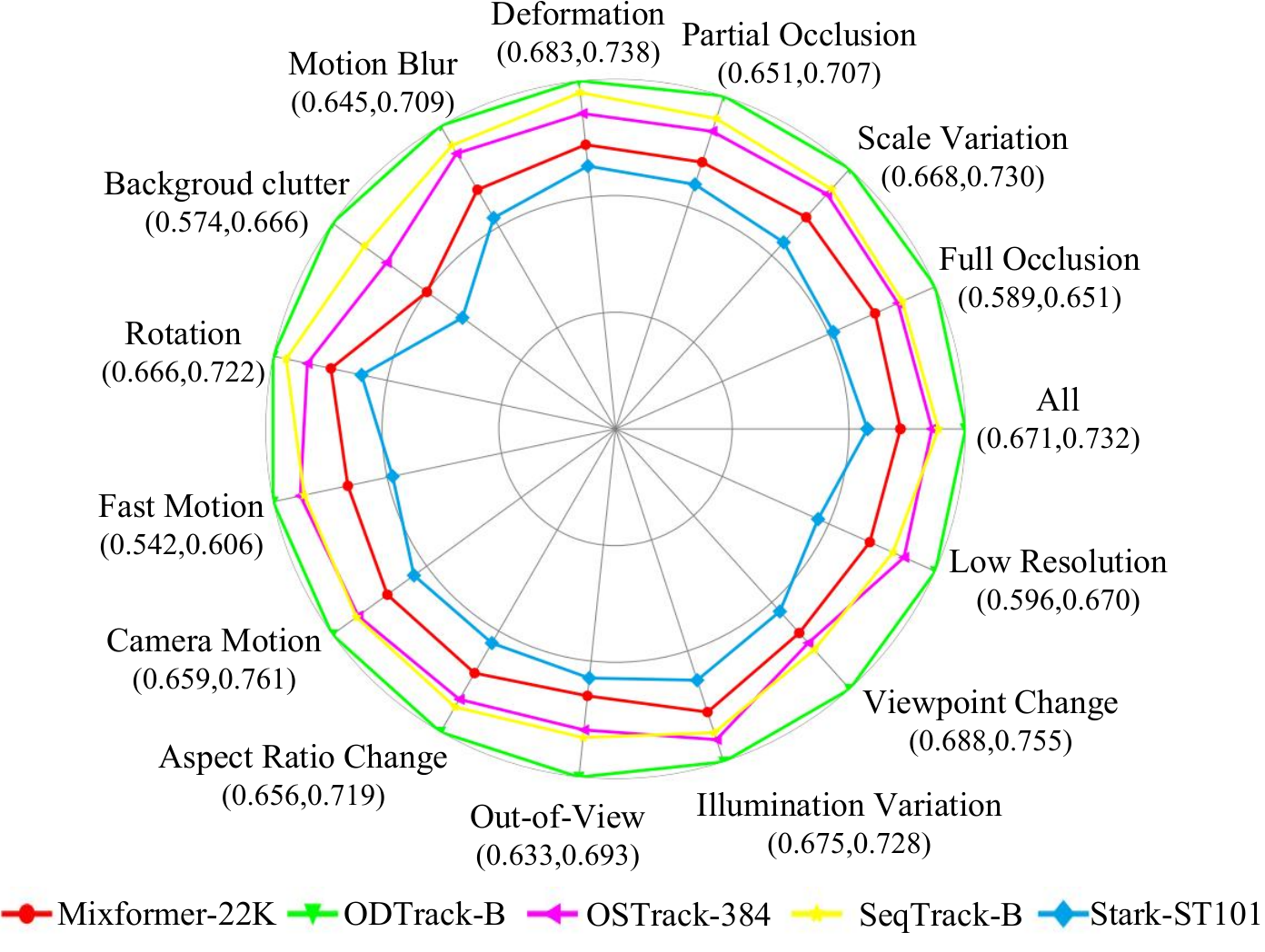}
       \caption{AUC scores of different attributes on LaSOT.}
       \label{fig:attrs}
    \end{figure}

\subsection{Implementation Details}

   \textbf{Training.}
   We use ViT-Base \cite{vit} model as the visual encoder, and its parameters are initialized with MAE\cite{MAE} pre-training parameters.
   The training data includes LaSOT \cite{lasot}, GOT-10k \cite{got10k}, TrackingNet \cite{trackingnet}, and COCO \cite{coco}.
   In terms of input data, we take the video sequence including three reference frames with $192 \times 192$ pixels and two search frames with $384 \times 384$ pixels as the input to the model.
   We employ the AdamW to optimize the network parameters with initial learning rate of $1 \times 10^{-5}$ for the backbone, $1 \times 10^{-4}$ for the rest, and set the weight decay to $10^{-4}$.
   We set the training epochs to 300 epochs. $60,000$ image pairs are randomly sampled in each epoch.
   The learning rate drops by a factor of 10 after 240 epochs.
   The model is conducted on a server with two 80GB Tesla A100 GPUs and set the batch size to be 8.
   
   \textbf{Inference.}
   To align with the training setting, we incorporate three reference frames at equal intervals into our tracker during the inference phase. Concurrently, search frames and temporal token vectors are input frame-by-frame.
   Further, we conduct comparative experiments in model parameters, FLOPs and inference speed, as shown in Tab.\ref{tab:param}.
   The proposed ODTrack is tested on a 2080Ti, and it runs at 32 $fps$.


\subsection{Comparison with the SOTA}

\begin{table*}[t]
    \centering
    \caption{Comparison with state-of-the-arts on four popular benchmarks: GOT10K, LaSOT, TrackingNet, and LaSOT$_{\rm{ext}}$. Where $*$ denotes for trackers only trained on GOT10K. The best two results are highlighted in {\color{red}red} and {\color{blue}blue}, respectively.}
    \label{tab:results}
    \resizebox{\textwidth}{!}{
    \begin{tabular}{l|ccc|ccc|ccc|ccc}
    \toprule
     \multicolumn{1}{c|}{\multirow{2}{*}{Method}} & \multicolumn{3}{c|}{GOT10K$^*$} & \multicolumn{3}{c|}{LaSOT} & \multicolumn{3}{c|}{TrackingNet} & \multicolumn{3}{c}{LaSOT$_{\rm{ext}}$} \\
     \cline{2-13}
      & AO & SR${_{0.5}}$ & SR${_{0.75}}$ & AUC & P${_{\rm{Norm}}}$ & P & AUC & P${_{\rm{Norm}}}$ & P & AUC & P${_{\rm{Norm}}}$ & P \\
      \midrule
      SiamFC \cite{SiamFC} & 34.8 & 35.3 & 9.8 & 33.6 & 42.0 & 33.9 & 57.1 & 66.3 & 53.3 & 23.0 & 31.1 & 26.9 \\
      ATOM \cite{atom} & 55.6 & 63.4 & 40.2 & 51.5 & 57.6 & 50.5 & 70.3 & 77.1 & 64.8 & 37.6 & 45.9 & 43.0 \\
      SiamPRN++ \cite{SiamRPN++} & 51.7 & 61.6 & 32.5 & 49.6 & 56.9 & 49.1 & 73.3 & 80.0 & 69.4 & 34.0 & 41.6 & 39.6 \\
      DiMP \cite{DiMP50} & 61.1 & 71.7 & 49.2 & 56.9 & 65.0 & 56.7 & 74.0 & 80.1 & 68.7 & 39.2 & 47.6 & 45.1 \\
      SiamRCNN \cite{siamrcnn} & 64.9 & 72.8 & 59.7 & 64.8 & 72.2 & - & 81.2 & 85.4 & 80.0 & - & - & - \\
      Ocean \cite{Ocean} & 61.1 & 72.1 & 47.3 & 56.0 & 65.1 & 56.6 & - & - & - & - & - & - \\
      STMTrack \cite{STMTrack} & 64.2 & 73.7 & 57.5 & 60.6 & 69.3 & 63.3 & 80.3 & 85.1 & 76.7 & - & - & - \\
      TrDiMP \cite{trdimp} & 67.1 & 77.7 & 58.3 & 63.9 & - & 61.4 & 78.4 & 83.3 & 73.1 & - & - & - \\
      TransT \cite{transt} & 67.1 & 76.8 & 60.9 & 64.9 & 73.8 & 69.0 & 81.4 & 86.7 & 80.3 & - & - & - \\
      Stark \cite{stark} & 68.8 & 78.1 & 64.1 & 67.1 & 77.0 & - & 82.0 & 86.9 & - & - & - & - \\
      SBT-B \cite{SBT} & 69.9 & 80.4 & 63.6 & 65.9 & - & 70.0 & - & - & - & - & - & - \\
      Mixformer \cite{mixformer} & 70.7 & 80.0 & 67.8 & 69.2 & 78.7 & 74.7 & 83.1 & 88.1 & 81.6 & - & - & - \\
      TransInMo \cite{TransInMo} & - & - & - & 65.7 & 76.0 & 70.7 & 81.7 & - & - & - & - & - \\
      OSTrack \cite{ostrack} & 73.7 & 83.2 & 70.8 & 71.1 & 81.1 & 77.6 & 83.9 & 88.5 & 83.2 & 50.5 & 61.3 & 57.6 \\
      AiATrack \cite{aiatrack} & 69.6 & 80.0 & 63.2 & 69.0 & 79.4 & 73.8 & 82.7 & 87.8 & 80.4 & 47.7 & 55.6 & 55.4 \\
      SeqTrack \cite{seqtrack} & 74.5 & 84.3 & 71.4 & 71.5 & 81.1 & 77.8 & 83.9 & 88.8 & 83.6 & 50.5 & 61.6 & 57.5 \\
      GRM \cite{GRM} & 73.4 & 82.9 & 70.4 & 69.9 & 79.3 & 75.8 & 84.0 & 88.7 & 83.3 & - & - & - \\
      VideoTrack \cite{VideoTrack} & 72.9 & 81.9 & 69.8 & 70.2 & - & 76.4 & 83.8 & 88.7 & 83.1 & - & - & - \\
      ARTrack \cite{ARTrack} & 75.5 & 84.3 & 74.3 & 72.6 & 81.7 & 79.1 & \color{blue}85.1 & 89.1 & 84.8 & 51.9 & 62.0 & 58.5 \\
      \hline
      \textbf{ODTrack-B} & \color{blue}77.0 & \color{red}87.9 & \color{blue}75.1 & \color{blue}73.2 & \color{blue}83.2 & \color{blue}80.6 & \color{blue}85.1 & \color{blue}90.1 & \color{blue}84.9 & \color{blue}52.4 & \color{blue}63.9 & \color{blue}60.1 \\
      \textbf{ODTrack-L} & \color{red}78.2& \color{blue}87.2 & \color{red}77.3 & \color{red}74.0 & \color{red}84.2 & \color{red}82.3 & \color{red}86.1 & \color{red}91.0 & \color{red}86.7 & \color{red}53.9 & \color{red}65.4 & \color{red}61.7 \\
    \bottomrule
    \end{tabular} }
\end{table*}

\textbf{GOT10K.}
GOT10K is a large-scale tracking dataset that contains more than 10,000 video sequences. The GOT10K benchmark proposes a protocol, which the trackers only use its training set for training. We follow the protocol to train our framework. As shown in Tab.\ref{tab:results}, the proposed method outperforms previous trackers and exhibits very competitive performance (77.0\% AO) when compared to the previous best-performing tracker ARTrack (75.5\% AO).
These results demonstrate that one benefit of our ODTrack comes from the video-level sample strategy, which is design to release the potential of video-level modeling framework.

\textbf{LaSOT.}
LaSOT is a large-scale long-term tracking benchmark that includes 1120 sequences for training and 280 sequences for testing. As shown in Tab.\ref{tab:results}, compared to most other tracking algorithms, our ODTrack-B achieves a new state-of-the-art result. For example, compared with the latest ARTrack, our method achieves 0.6\%, 1.5\%, and 1.5\% gains in terms of AUC, P${_{\rm{Norm}}}$ and P score, respectively. 
Furthermore, Fig.\ref{fig:attrs} shows the results of attribute evaluation, demonstrating that our tracker outperforms other tracking methods on multiple challenge attributes.
These results show that the token propagation mechanism helps the model to learn trajectory information about the target instance and greatly improves target localization in long-term tracking scenarios.

\textbf{TrackingNet.}
TrackingNet is a large-scale short-term dataset that provides a test set with 511 video sequences.
As reported in Tab.\ref{tab:results}, compared with the high-preformance tracker SeqTrack, our method achieves good tracking results that outperform 1.2\%, 1.3\%, and 1.3\% in terms of success, normalized precision and precision score, respectively.
This demonstrates that our ODTrack exhibits strong generalization capabilities.

\textbf{LaSOT$_{\rm{ext}}$.}
LaSOT$_{\rm{ext}}$ is the extended version of LaSOT, which comprises 150 long-term video sequences.
As reported in Tab.\ref{tab:results}, our method achieves the good tracking results that outperform most compared trackers. For example, our tracker gets a AUC of 52.4\%, $P_{Norm}$ score of 63.9\%, and $P$ score of 60.1\%, outperforming the ARTrack by 0.5\%, 1.9\%, and 1.6\%, respectively.
There results meet our expectation that video-level modeling has more stable object localization capabilities in complex scenarios.

\textbf{VOT2020.}
VOT2020\cite{VOT2020} contains 60 challenging sequences, and it uses binary segmentation masks as the groundtruth. We use Alpha-Refine \cite{Alpha-Refine} as a post-processing network for ODTrack to predict segmentation masks.
As shown in Tab.\ref{tab:vot2020}, our ODTrack-B and -L achieve the best results with EAO of 58.1\% and 60.5\% on mask evaluations, respectively. 

\textbf{TNL2K and OTB100.}
We evaluate our tracker on TNL2K\cite{tnl2k} and OTB100\cite{OTB2015} benchmarks. They include 700 and 100 video sequences, respectively. These results in Tab.\ref{tab:tnl2k} show that the ODTrack-B and -L achieve the best performance on TNL2K and OTB100 benchmarks, demonstrating the effectiveness of the temporal token propagation attention mechanism.

\begin{table}[t]
\centering
\caption{State-of-the-art comparison on VOT2020.}
\label{tab:vot2020}
\resizebox{\linewidth}{!}{
\begin{tabular}{l|ccc}
\toprule
Method & EAO $(\uparrow)$ & Accuracy $(\uparrow)$ & Robustness $(\uparrow)$ \\
\midrule
SiamMask & 0.321 & 0.624 & 0.648 \\
Ocean & 0.430 & 0.693 & 0.754 \\
D3S & 0.439 & 0.699 & 0.769 \\
SuperDiMP & 0.305 & 0.492 & 0.745 \\
AlphaRef & 0.482 & 0.754 & 0.777 \\
STARK & 0.505 & 0.759 & 0.819 \\
SBT & 0.515 & 0.752 & 0.825 \\
Mixformer & 0.535 & {\color{blue}0.761} & 0.854 \\
SeqTrack-B & 0.522 & - & - \\
\hline
\textbf{ODTrack-B} & {\color{blue}0.581} & {\color{red}0.764} & {\color{blue}0.877} \\
\textbf{ODTrack-L} & {\color{red} 0.605} & {\color{blue}0.761} & {\color{red} 0.902} \\
\bottomrule
\end{tabular}}
\end{table}


\begin{table*}
  \centering
  \caption{Ablation Studies of different token propagation designs on LaSOT benchmark.}
  \label{tab:study}
\begin{subtable}[t]{0.3\textwidth}
\footnotesize
  \centering
\caption{Comparison on propagation method}
  \begin{tabular}{l|ccc}
\toprule
Method & AUC & $P_{Norm}$ & $P$ \\
\midrule
\textit{Baseline} & 70.1 & 80.2 & 76.9 \\
$w/o$ \textit{Token} & 71.0 & 81.1 & 78.0 \\
\textit{Separate} & 72.2 & 82.3 & 79.2 \\
\textit{Concatenation} & {\color{red} 72.8} & {\color{red} 83.0} & {\color{red} 80.3} \\
\bottomrule
\end{tabular}
\end{subtable}
  \hfill
\begin{subtable}[t]{0.3\textwidth}
\footnotesize
  \centering
\caption{Comparison on video sequence length}
\begin{tabular}{c|ccc}
\toprule
Sequence Length & AUC & $P_{Norm}$ & $P$ \\
\midrule
2 & 72.8 & 83.0 & 80.3 \\
3 & \color{red}73.1 & \color{red}83.0 & \color{red}80.4 \\
4 & 72.5 & 82.9 & 79.9 \\
5 & 72.0 & 82.1 & 79.3 \\
\bottomrule
\end{tabular}
\end{subtable}
  \hfill
\begin{subtable}[t]{0.3\textwidth}
  \centering
  \footnotesize
\caption{Comparison on sampling range}
  \begin{tabular}{c|ccc}
\toprule
Sample Range & AUC & $P_{Norm}$ & $P$ \\
\midrule
200 & 72.8 & 83.0 & 80.3 \\
400  & {\color{red} 73.1} & {\color{red} 83.5} & {\color{red} 80.6} \\
800  & 73.0 & 83.3 & 80.4 \\
1200  & 73.0 & 83.1 & 80.1 \\
\bottomrule
\end{tabular}
\end{subtable}
\end{table*}

\begin{table*}[t]
\centering
\caption{Comparison with state-of-the-art methods on TNL2K and OTB100 benchmarks in AUC score.}
\label{tab:tnl2k}
\resizebox{\textwidth}{!}{
\begin{tabular}{c|cccccccccc|cc}
\toprule
& ATOM & Ocean & DiMP & TransT & TransInMo & OSTrack & SBT & Mixformer & SeqTrack-B & ARTrack & \textbf{ODTrack-B} & \textbf{ODTrack-L} \\
\midrule
TNL2K & 40.1 & 38.4 & 44.7 & 50.7 & 52.0 & 55.9 & - & - & 56.4 & 59.8 & \color{blue}60.9 & \color{red}61.7 \\
OTB100 & 66.3 & 68.4 & 68.4 & 69.6 & 71.1 & - & 70.9 & 70.0 & - & - & \color{blue}72.3 & \color{red}72.4 \\
\bottomrule
\end{tabular} }
\end{table*}


\subsection{Ablation Study}

\textbf{Importance of token propagation.}
To investigate the effect of token propagation in Eq.\ref{eq:propagate}, we perform experiments whether propagating temporal token in Tab.\ref{tab:study}(a).
$w/o$ \textit{Token} denotes the experiment employing video-level sampling strategy without token propagation.
From the second and third rows, it can be observed that the absence of the token propagation mechanism leads to a decrease in the AUC score by 1.2\%.
This result indicates that token propagation plays a crucial role in cross-frame target association.

\textbf{Different token propagation methods.}
We conduct experiments to validate the effectiveness of two proposed token propagation methods in the video-level tracking framework in Tab.\ref{tab:study}(a). We can be observe that both the \textit{separate} and \textit{concatenation} methods achieve significant performance improvements, with the \textit{concatenation} method showing slightly better results. This demonstrates the effectiveness of both attention mechanisms.

\textbf{The length of search video-clip.}
As shown in Tab.\ref{tab:study}(b), we ablate the impact of search video sequence length on the tracking performance. 
When the length of video clip increases from 2 to 3, the AUC metric improves by 0.3\%. However, continuous increment in sequence length does not lead to performance improvement, indicating that overly long search video clips impose a learning burden on the model. Hence, we should opt for an appropriate the length of search video clip.

\textbf{The sampling range.}
To validate the impact of sampling range on algorithm performance, we conduct experiments on the sampling range of video frames in Tab.\ref{tab:study}(c).
When the sampling range is expanded from 200 to 1200, there is a noticeable improvement in performance on the AUC metric, indicating that the video-level framework can learn target trajectory information from a larger sampling range.

   \begin{figure}[t]
      \centering
      \includegraphics[width=1\linewidth]{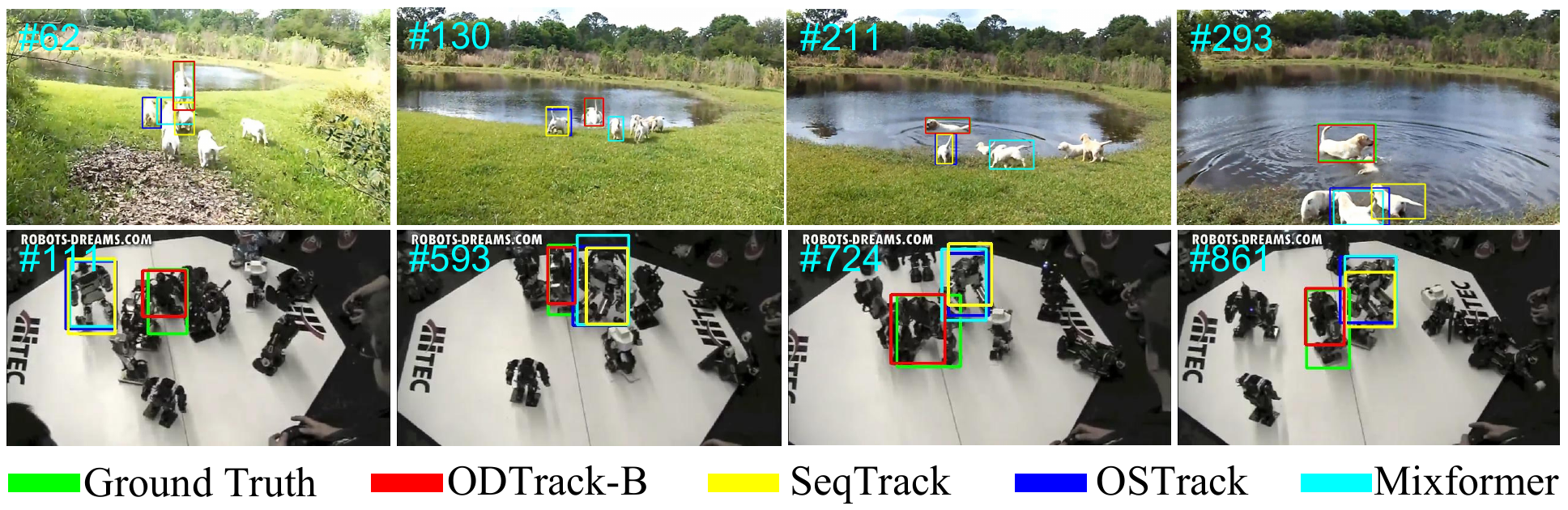}
       \caption{Qualitative comparison results of our tracker with other three SOTA trackers on LaSOT benchmark.
       }
       \label{fig:visual}
    \end{figure}

   \begin{figure}[t]
      \centering
      \includegraphics[width=1\linewidth]{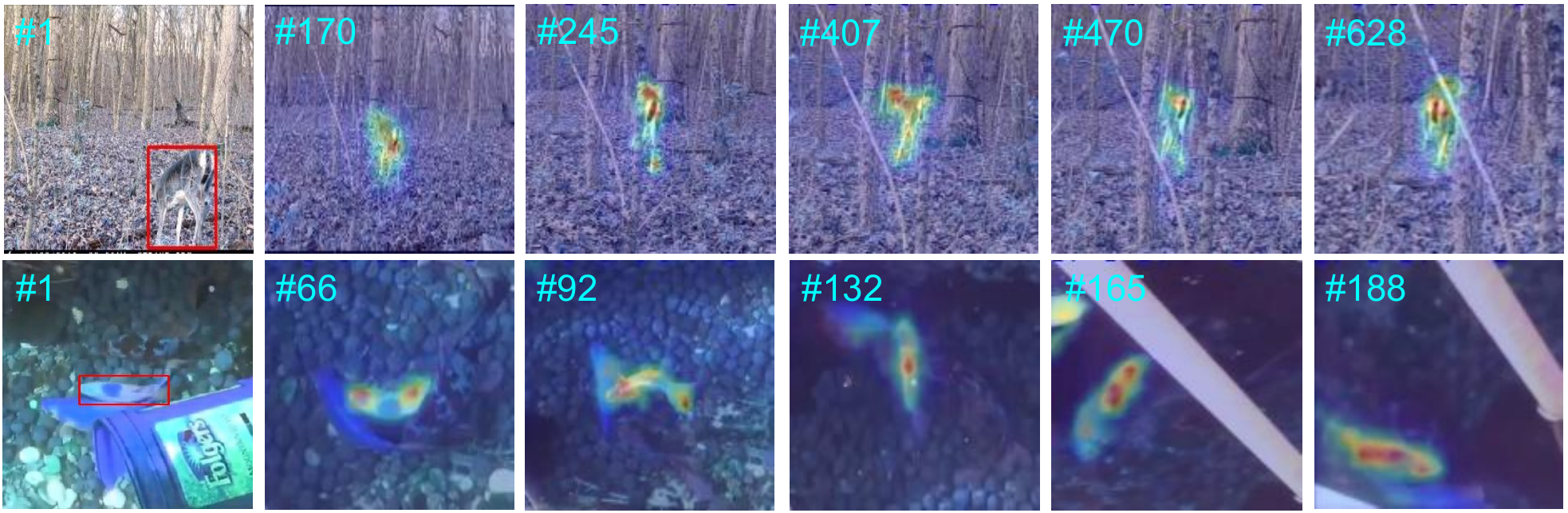}
       \caption{The attention map of temporal token attention operation.}
       \label{fig:attnmap}
    \end{figure}

\subsection{Visualization and Limitation}


\textbf{Visualization.}
To intuitively show the effectiveness of the proposed method, especially in complex scenarios including similar distractors, we visualize the tracking results of our ODTrack and three advanced trackers on LaSOT dataset. As shown in Fig.\ref{fig:visual}, due to its ability to densely propagate trajectory information of the target, our tracker far outperforms the latest tracker SeqTrack on these sequences.

Furthermore, we visualize the attention map of temporal token attention operation, as shown in Fig.\ref{fig:attnmap}.
We can observe that the temporal token continuously propagate and attend to motion trajectory information of object, which aids our tracker in accurately localizing target instance.

\textbf{Limitation.}
This work models the entire video as a sequence and decode the localization of instance frame by frame in an auto-regressive manner. Despite achieving remarkable results, our video-level modeling method is a global approximation due to constraints in GPU resources, and we are still unable to construct the framework in a cost-effective manner. A promising solution would involve improving the computational complexity and lightweight modeling of the transformer.

\section{Conclusion}
In this work, we present ODTrack, a new video-level framework for visual object tracking. We reformulate visual tracking as a token propagation task that densely associates the contextual relationships of across video frames in an auto-regressive manner.
Furthermore, we propose a video sequence sampling strategy and two temporal token propagation attention mechanisms, enabling the proposed framework to simplify video-level spatio-temporal modeling and avoid intricate online update strategies.
Extensive experiments show that our ODTrack achieves promising results on seven tracking benchmarks. We hope that this work inspires further research in video-level tracking modeling.

\section{Acknowledgements}
This work is supported by the National Natural Science Foundation of China (No.U23A20383, 61972167 and U21A20474), the Project of Guangxi Science and Technology (No.2022GXNSFDA035079 and 2023GXNSFDA026003), the Guangxi "Bagui Scholar" Teams for Innovation and Research Project, the Guangxi Collaborative Innovation Center of Multi-source Information Integration and Intelligent Processing, the Guangxi Talent Highland Project of Big Data Intelligence and Application, and the Research Project of Guangxi Normal University (No.2022TD002).

\bibliography{aaai24}

\end{document}